\documentclass[runningheads]{llncs}
\usepackage{graphicx}
\usepackage[breaklinks=true,bookmarks=false]{hyperref}
\begin{document}
\title{An Open-source Tool for Hyperspectral Image Augmentation in Tensorflow}
%
%
\author{Mohamed Abdelhack\inst{1}\orcidID{0000-0002-6753-3237}}
\authorrunning{M. Abdelhack}
%
\institute{Washington University in St. Louis, St. Louis MO, USA
\email{mohamed.abdelhack.37a@kyoto-u.jp}}
\maketitle              
\begin{abstract}
   Satellite imagery allows a plethora of applications ranging from weather forecasting to land surveying. The rapid development of computer vision systems could open new horizons to the utilization of satellite data due to the abundance of large volumes of data. However, current state-of-the-art computer vision systems mainly cater to applications that mainly involve natural images. While useful, those images exhibit a different distribution from satellite images in addition to having more spectral channels. This allows the use of pretrained deep learning models only in a subset of spectral channels that are equivalent to natural images thus discarding valuable information from other spectral channels. This calls for research effort to optimize deep learning models for satellite imagery to enable the assessment of their utility in the domain of remote sensing. Tensorflow tool allows for rapid prototyping and testing of deep learning models, however, its built-in image generator is designed to handle a maximum of four spectral channels. This manuscript introduces an open-source tool that allows the implementation of image augmentation for hyperspectral images in Tensorflow. Given how accessible and easy-to-use Tensorflow is, this tool would provide many researchers with the means to implement, test, and deploy deep learning models for remote sensing applications.

\end{abstract}
\section{Introduction}
Remote sensing applications have recently been utilizing deep learning techniques with an increasing pace \cite{ma2019}. However, tools for utilization of of deep neural networks (DNN) are still not well-developed for the field of satellite imagery where most of the methods are imported as is from natural image processing that are restricted to the visible range of the electromagnetic spectrum. Satellite images are usually hyperspectral and they also capture images with statistics different to natural images. One of the techniques widely used in deep learning is image augmentation. It have been a very successful technique to increase the variety of natural images used for training deep neural networks \cite{cirecsan2011,ciregan2012,simard2003}. The advantage of this technique is that it enlarges the training dataset while preserving labels by utilizing symmetries in natural images.
While there are multiple studies that managed to utilize image augmentation in training deep learning models on hyperspectral images \cite{scott2017,qiu2019}, they usually went the route of creating their own tools. The lack of a widely available, open-source tool could be an obstacle for many researchers in the field. This paper introduces a tool that could be integrated with Tensorflow tool cite{tensorflow2015} which is a widely-used tool for deep learning prototyping and research. Its ease of use has allowed a wide audience to explore the world of deep learning and develop their own applications without an extensive knowledge of programming. One drawback of Tensorflow is that its native image augmentation tool only allows up to four-channel images. The tool introduced in this paper allows this audience to utilize full Tensorflow functionality along with image augmentation tools for hyperspectral images. Making such a tool widely available could assist in improving the DNN models for satellite images which could open new horizons in the field of remote sensing especially with the high volume of satellite images that cover the whole Earth. This could enable many researchers especially in developing areas to make use of the open access satellite data currently available through portals such as the Copernicus Open Access Hub.
\section{Methods}
\subsection{Data Augmentation Tool}
Due to the lack of a Keras/Tensorflow implementation of the image generator tool that supports hyperspectral images, I developed an image generator seeding from previously published codes \cite{leitloff2018} but with extended functionality using Python Scikit-image toolbox. The tool allows for image augmentation by horizontal and vertical flipping, rotation, translation, zooming, shearing, and addition of speckle (multiplicative) noise. All but the latter technique are common procedures for natural image augmentation in training DNN models. Speckle noise functionality addition was added as an augmentation technique that better simulates the types of noise that affect satellite images \cite{vijay2012}. The toolbox is available on GitHub\footnote{\url{https://github.com/mabdelhack/hyperspectral_image_generator}.} under an MIT License.\\
Additionally, to allow image generation from satellite images along with shape files that specify data points, I implemented a function where the input to the generator is a JPEG 2000 format images, a common format for satellite images, and a shape file that comprises of coordiantes for the centroids of the map locations from where images need to be extracted. The user would specify the size of images as well as augmentation techniques to use and the tool would crop those images as the DNN is being trained (Figure 1).

\subsection{Model Architecture}
In order to make a simple test case, I trained a standard convolutional neural network VGG19 \cite{simonyan2014} as a base model for training. The input layer dimensions were expanded to incorporate thirteen channels of the hyperspectral satellite images. Additionally, the fully connected layers were replaced by three layers of dimension $2048 \times 2048$, $2048 \times 1024$, and $1024 \times 10$ where the last ten units correspond to the classification targets as will be further elaborated in the dataset description section. The model was implemented using Tensorflow with Keras implementation \cite{chollet2015}.
\begin{figure}
\includegraphics[width={\textwidth}]{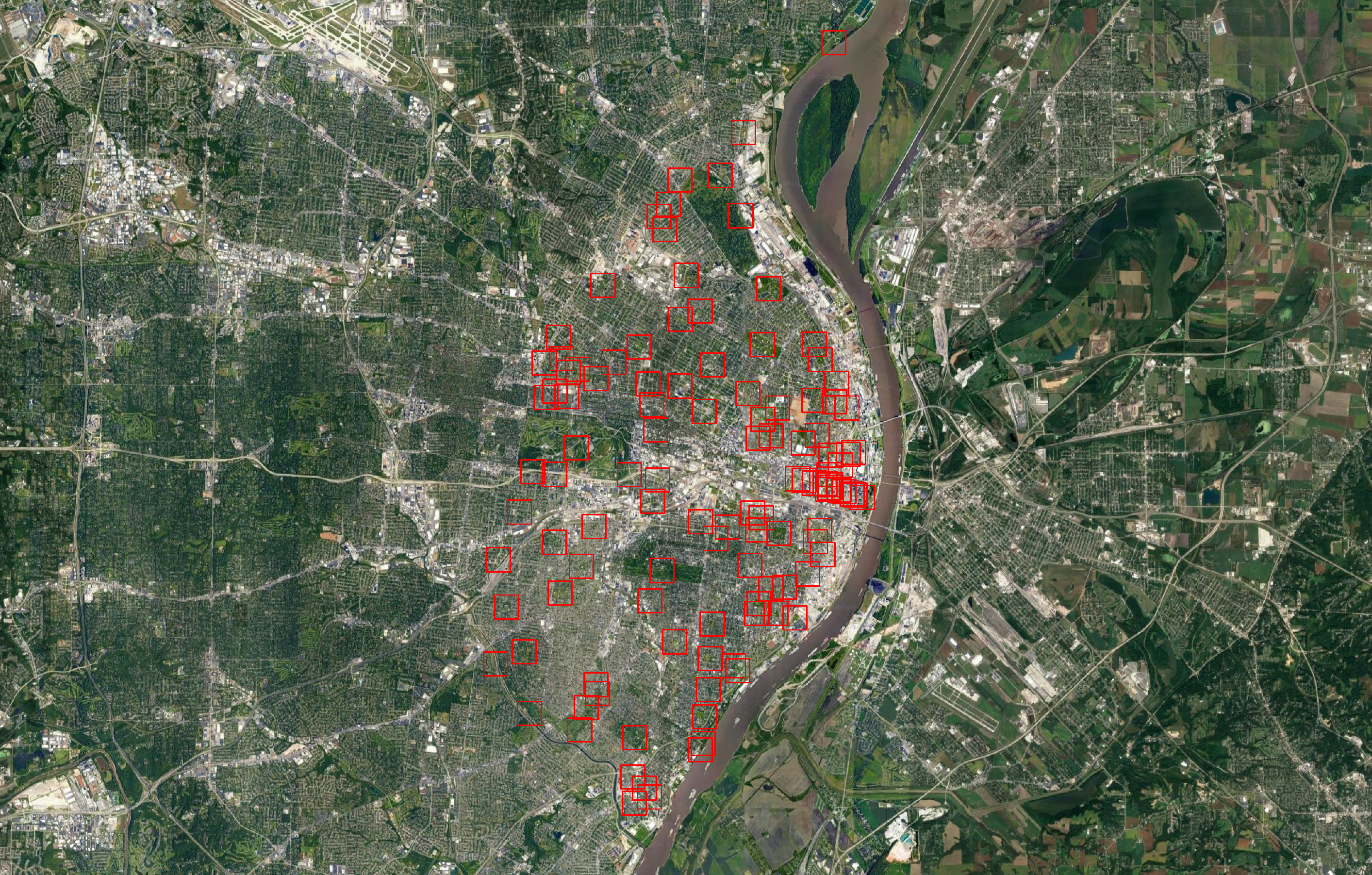}
\caption{A demonstration of direct image extraction from satellite images by using point coordinates. This example data shows parks in St. Louis City, MO extracted from \url{https://www.stlouis-mo.gov/} and base map retrieved from \url{maps.google.com} on 2020-July-8.} \label{fig1}
\end{figure}
\subsection{Dataset}
Hyperspectral images from EuroSAT were used \cite{helber2019}. Images in this dataset were taken by the satellite Sentinel-2A covering areas from different regions in Europe. Each image was 64 $\times$ 64 pixels with thirteen channels representing different electromagnetic bands. Each image belonged to one of ten classification targets. A more complete description of the data acquisition can be found at Helber et al. 2019 \cite{helber2019}.

\subsection{Data Augmentation Testing}
I first tested each of the common image augmentation techniques separately and performance was compared to the no augmentation condition. I tested horizontal and vertical flipping, rotation with a maximum range of $90^{\circ}$, zooming (in-out) with a maximum factor of $1.5$, translation with a maximum range of $25\%$ of image size, shearing with a maximum range of $5\%$. When augmentation causes image to go out of frame, the edge pixels are repeated to pad the missing regions and preserve the image size.
Secondly, I tested the introduction of speckle noise which is a multiplicative noise component common in satellite communication \cite{vijay2012} as an augmentation technique. I tested the addition of speckle noise with zero mean and variance of 0.010.

\subsection{Training and Performance Measurement}
I trained each DNN model for ten epochs with 500 batches per epoch of 128 images each using an Adam optimizer (learning rate=0.001) with a categorical cross-entropy loss function. Training dataset included 21000 training samples and 9000 test samples. This meant that without augmentation, each training sample would be used on average 30 times throughout the ten epochs of training which could lead to over-fitting given that the network has a total of 21,025,226 trainable parameters.

\section{Results}
Table 1 summarizes accuracy performance for the epoch with highest test accuracy for each of the different testing conditions. Results showed all augmentation techniques led to increased accuracy except for image translation. One reason for that decrease in performance could be the image padding mechanism that follows translation that could be destructive to the image. For this reason, an extension to the tool that extract images directly from the high resolution satellite images was included (Figure 1), This could improve the utility of translation augmentation if it were implemented on a larger tile where the image translation simply includes surrounding pixels from the larger tile rather than removing them completely and padding them later. Speckle noise augmentation technique also appears to be a promising approach to image augmentation for satellite data \cite{kwak2018}. Also, being native to satellite images, it could prove beneficial for generalization of the models to images originating from different satellites. These results show that traditional image augmentation techniques could improve performance even in hyperspectral setting as demonstrated by previous studies \cite{scott2017,qiu2019}.\\
\begin{table*}
\begin{center}
\begin{tabular}{|c|c|c|c|c|c|c|c|}
\hline
Augmentation Technique & Best Epoch & Training Accuracy & Test Accuracy\\
\hline\hline
No Augmentation & $7$ & $86.72\%$ & $87.66\%$\\
Horz/Vert Flipping & $9$ & $92.79\%$ & $93.43\%$\\
1.5 zooming & $9$ & $89.16\%$ & $88.59\%$\\
25\% translation & $7$ & $66.05\%$ & $70.00\%$\\
$90^{\circ}$ rotation & $9$ & $83.76\%$ & $90.00\%$\\
5\% shearing & $7$ & $93.99\%$ & $90.93\%$\\
Speckle Noise (Variance=0.010) & $8$ & $92.18\%$ & $91.25\%$\\
\hline
\end{tabular}
\end{center}
\caption{Results of training the DNN using image augmentation techniques.}
\end{table*}
\section{Conclusion}
Many of the traditional image augmentation techniques appear to be readily transferable to satellite image processing model, more specific augmentation methods could also further enhance the accuracy and generalizability of these models. Having a publicly available tool for image augmentation could assist many researchers in the field of remote sensing.


%
%
\bibliographystyle{splncs04}

\begin{thebibliography}{8}

\bibitem{tensorflow2015}
Mart\'{\i}n Abadi et al.,
\newblock {TensorFlow}: Large-scale machine learning on heterogeneous systems,
  2015.
\newblock Software available from tensorflow.org.

\bibitem{chollet2015}
François Chollet.
\newblock Keras.
\newblock \url{https://github.com/fchollet/keras}, 2015.

\bibitem{ciregan2012}
Dan Ciregan, Ueli Meier, and J{\"u}rgen Schmidhuber.
\newblock Multi-column deep neural networks for image classification.
\newblock In {\em 2012 IEEE conference on computer vision and pattern
  recognition}, pages 3642--3649. IEEE, 2012.


\bibitem{cirecsan2011}
Dan~C Cire{\c{s}}an, Ueli Meier, Jonathan Masci, Luca~M Gambardella, and
  J{\"u}rgen Schmidhuber.
\newblock High-performance neural networks for visual object classification.
\newblock {\em arXiv preprint arXiv:1102.0183}, 2011.


\bibitem{helber2019}
Patrick Helber, Benjamin Bischke, Andreas Dengel, and Damian Borth.
\newblock Eurosat: A novel dataset and deep learning benchmark for land use and
  land cover classification.
\newblock {\em IEEE Journal of Selected Topics in Applied Earth Observations
  and Remote Sensing}, 12(7):2217--2226, 2019.



\bibitem{kwak2018}
Youngchul Kwak, Woo-Jin Song, and Seong-Eun Kim.
\newblock Speckle-noise-invariant convolutional neural network for sar target
  recognition.
\newblock {\em IEEE Geoscience and Remote Sensing Letters}, 16(4):549--553,
  2018.

\bibitem{leitloff2018}
Jens Leitloff and Felix~M. Riese.
\newblock {Examples for CNN training and classification on Sentinel-2 data}.
\newblock
  \href{http://doi.org/10.5281/zenodo.3268451}{http://doi.org/10.5281/zenodo.3268451},
  2018.

\bibitem{ma2019}
Lei Ma, Yu Liu, Xueliang Zhang, Yuanxin Ye, Gaofei Yin, and Brian~Alan Johnson.
\newblock Deep learning in remote sensing applications: A meta-analysis and
  review.
\newblock {\em ISPRS journal of photogrammetry and remote sensing},
  152:166--177, 2019.

\bibitem{qiu2019}
Chunping Qiu, Lichao Mou, Michael Schmitt, and Xiao~Xiang Zhu.
\newblock Local climate zone-based urban land cover classification from
  multi-seasonal sentinel-2 images with a recurrent residual network.
\newblock {\em ISPRS Journal of Photogrammetry and Remote Sensing},
  154:151--162, 2019.

\bibitem{scott2017}
G.~J. {Scott}, M.~R. {England}, W.~A. {Starms}, R.~A. {Marcum}, and C.~H.
  {Davis}.
\newblock Training deep convolutional neural networks for land–cover
  classification of high-resolution imagery.
\newblock {\em IEEE Geoscience and Remote Sensing Letters}, 14(4):549--553,
  2017.

\bibitem{simard2003}
Patrice~Y Simard, David Steinkraus, John~C Platt, et~al.
\newblock Best practices for convolutional neural networks applied to visual
  document analysis.
\newblock In {\em Icdar}, volume~3, 2003.

\bibitem{simonyan2014}
Karen Simonyan and Andrew Zisserman.
\newblock Very deep convolutional networks for large-scale image recognition.
\newblock {\em arXiv preprint arXiv:1409.1556}, 2014.

\bibitem{vijay2012}
M Vijay and L~Saranya Devi.
\newblock Speckle noise reduction in satellite images using spatially adaptive
  wavelet thresholding.
\newblock {\em International Journal of Computer Science and Information
  Technologies (IJCSIT)}, 3(2012):3432--3435, 2012.

\end{thebibliography}
%

\end{document}